\documentclass[runningheads]{llncs}
\usepackage{graphicx}
\usepackage{paralist}
\usepackage{multirow}
\usepackage{tabularx}
\usepackage{amsmath}
\usepackage{amssymb}
\usepackage{comment}
\usepackage{array}

\usepackage[breaklinks=true,
    colorlinks=true,
    linkcolor=black,
    citecolor=black,
    urlcolor=black]{hyperref}

\renewcommand{\thepage}{\arabic{page}}
\def\hb{\hbox to 11.5 cm{}}
\begin{document}
\begin{frontmatter}
\title{OntoPret: An Ontology for the Interpretation of Human Behavior}
\author{Alexis Ellis\inst{1} \and Stacie Severyn\inst{1} \and Fjollë Novakazi\inst{2} \and Hadi Banaee\inst{2} \and Cogan Shimizu\inst{1}}
\institute{Wright State University, USA\\\email{ellis.177@wright.edu} \and Örebro University, SE}
\authorrunning{Ellis, A., et al.}
\end{frontmatter}
\markboth{April 2025\hb}{April 2025\hb}
\maketitle
\begin{abstract}
  As human-machine teaming becomes central to paradigms like Industry 5.0, a critical need arises for machines to safely and effectively interpret complex human behaviors. A research gap currently exists between techno-centric robotic frameworks, which often lack nuanced models of human behavior, and descriptive behavioral ontologies, which are not designed for real-time, collaborative interpretation. This paper addresses this gap by presenting OntoPret, an ontology for the interpretation of human behavior. Grounded in cognitive science and a modular engineering methodology, OntoPret provides a formal, machine-processable framework for classifying behaviors, including task deviations and deceptive actions. We demonstrate its adaptability across two distinct use cases—manufacturing and gameplay—and establish the semantic foundation necessary for advanced reasoning about human intentions.
\end{abstract}
\section{Introduction}
\label{sec:intro}
As Industry 4.0 evolves towards the more human-centric paradigm of Industry 5.0 \cite{5.0industry1,industry4.0}, intelligent agents are increasingly integrated into collaborative workspaces, creating new challenges for human-machine interaction \cite{aiindustrial,smartrobotic}. While research has advanced human-machine co-working \cite{cowork}, a fundamental complexity remains: human behavior itself \cite{valcri}. Unlike programmable tasks, human actions are nuanced and context-dependent, often leading to unexpected deviations or even deceptive behavior. Effectively supporting true human-centric teaming, therefore, requires machines to move beyond simple task execution and develop a structured, interpretable understanding of their human counterparts.

To address this, researchers have developed robotics-centered frameworks like KnowRob, which provides detailed knowledge for robotic tasks but lacks effective integration of human collaborators \cite{beetz2018know}. Similarly, platforms such as ORO support interaction in human-inhabited domains but do not model the specific context of human intentions \cite{lemaignan2010oro}. While other approaches integrate aspects like contextual awareness \cite{umbrico2020ontology} or intention recognition \cite{hadi}, they often lack a deep, formal model of the nuanced behavioral patterns of the human agent.

Conversely, ontologies focused on human behavior, such as the Human Behavior Ontology (HBO) \cite{human} and the Behavior Change Technique Ontology (BCTO) \cite{marques2024behaviour,michie2021representation}, offer structured classifications of human actions. 

However, while valuable for behavioral analysis, these ontologies lack the specific focus on real-time, collaborative, and contextual interpretation required for human-machine interaction scenarios.

\subsection{Aim and Contributions}
The aforementioned challenges reveal a critical research gap: a need for a framework that bridges the divide between techno-centric robotic systems and descriptive behavioral models. Such a framework must be grounded in cognitive science to formally represent the context, intentions, and variations in human actions, enabling safer and more resilient human-machine collaboration.

This paper addresses this gap by introducing the Ontology for the Interpretation of Human Behavior (OntoPret). Our aim is to provide a general, human-centric framework that allows machines to engage with humans more effectively by interpreting complex behaviors like deviations and deception. To achieve this, we utilize a modular ontology design grounded in cognitive science to enable systematic behavior interpretation. We demonstrate the framework's adaptability by applying it to two distinct use cases - manufacturing and gameplay - and establish the semantic foundation necessary for advanced reasoning tasks, such as real-time intention recognition. OntoPret is openly available online \cite{anonymous}.

\section{Foundational Concepts}
\label{sec:back}
We distinguish between \textbf{interpretation} - the ontological structuring and classification of observed behaviors - and \textbf{reasoning}, the computational processes that utilize this knowledge. This distinction is critical, as OntoPret’s primary contribution is providing the formal framework for interpretation, upon which advanced reasoning about deviations and intentions can be built.

\subsection{Core Foundational Concepts}
\label{ssec:dec}
\begin{figure}[t]
    \centering
    \includegraphics[width=1\textwidth]{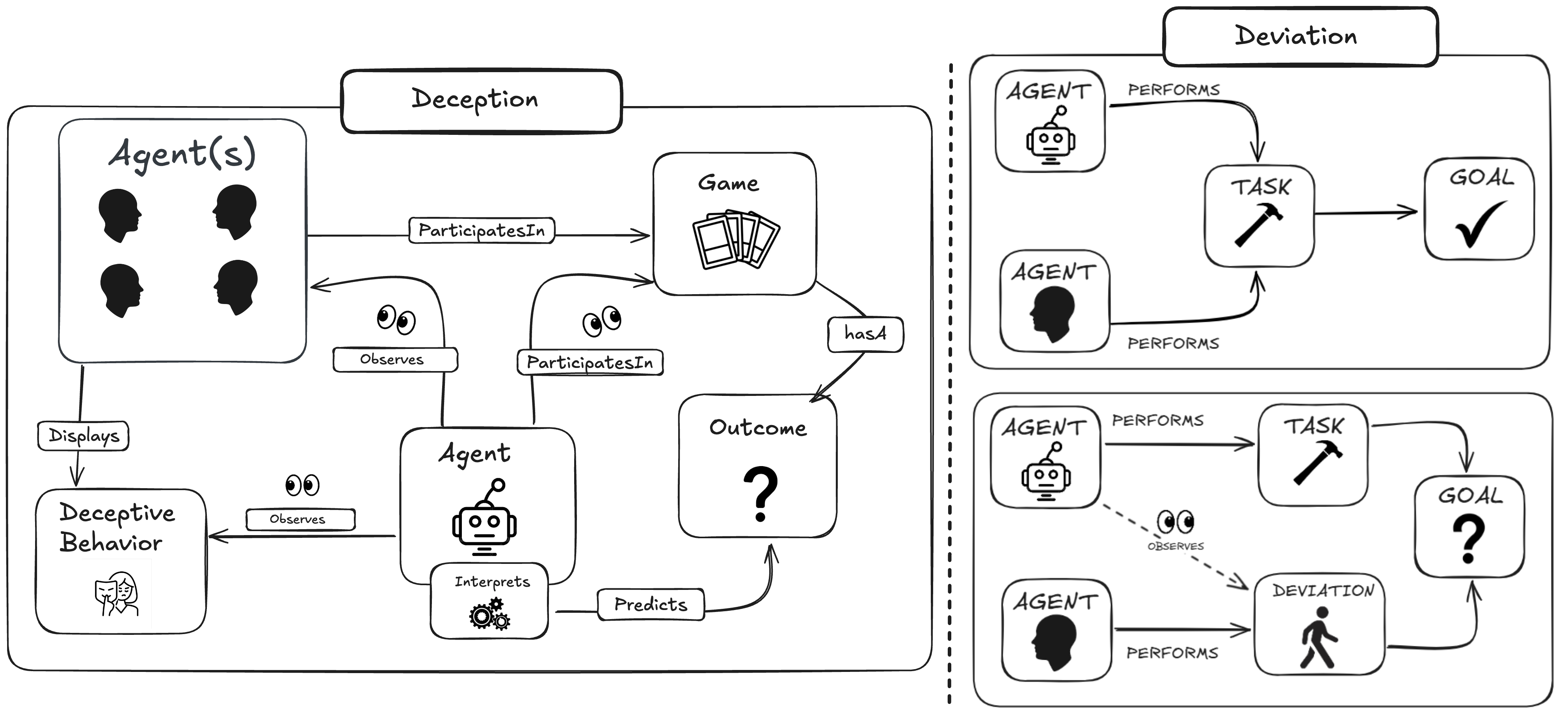}
    \caption{Simple representations of Deception and Deviation}
    \label{fig:breakdown}
\end{figure}
\paragraph{Deviation}
In human-machine interaction, a \textbf{deviation} is an action that diverges from an expected or prescribed sequence, whether as a violation of human anticipation \cite{dev1}, a compromise of task success under industrial standards \cite{dev2}, or a mismatch in system behavior requiring adaptation \cite{dev3}. 
For instance, if a human collaborator on an assembly line unexpectedly walks away from their task, the robotic agent must recognize this deviation and interpret its potential impact on the shared goal. 
As in Figure~\ref{fig:breakdown}, the ability to assess and reason about such deviations is a critical step towards creating safer and more efficient collaborative workspaces.

\paragraph{Deception}
For this paper, we define \textbf{deception} as an agent causing a target agent to believe something they would not otherwise believe \cite{chisholm_intent_1977,mahon_definition_2015,sarkadi_deception_2024}. The relevance of interpreting such behavior is clear in scenarios like a poker game, as depicted in Figure~\ref{fig:breakdown}. Unlike games of perfect information (e.g., chess, Go, or checkers), winning at poker often requires interpreting an opponent's intentionally misleading behaviors, or ``tells,'' to assess their hand strength \cite{elwood_reading_2012,guazzini_emergence_2009}. A framework that facilitates this interpretation can thus create more realistic human-machine gameplay. 

Machines with the ability to autonomously detect deception could provide an important aid to human-human and human-machine interactions, with the caveat that it is necessary to consider an ethical perspective. Deception can be found within areas such as criminal investigations, military strategies, or personal relationships. 
We acknowledge, however, that care should be taken that the impacts of the technology on human rights and well-being are considered \cite{king_applications_2024}.

\subsection{Cognitive Foundations for Behavior Interpretation}
\label{sssec:ToM-MM}
Effective human-machine teaming aims to combine the strengths of humans and intelligent systems to enhance productivity and safety \cite{Li_2024}. Achieving this, particularly within the human-centric paradigm of Industry 5.0 \cite{Nahavandi2019Industry50,Rahman2024Cobotics}, requires machines to put the human at the center and therefore, to understand human intentions and adapt to their needs \cite{Ottoni_2024}. 

A key aspect of this is the system's ability to develop a ``Theory of Mind'' (ToM)---the ability to infer and understand human mental states such as beliefs, intentions, and goals. This capability allows robots to anticipate human actions, recognize their goals, and respond appropriately in collaborative tasks.

Closely related are mental models, which are an agent's internal representation of its own beliefs, intentions, and goals \cite{gentner2014mental}, helping it to plan future actions \cite{westbrook2006mental}. ToM and mental models are foundational components in human-machine collaboration, enabling systems to build internal models of human behavior to support coordination and adaptability \cite{carroll1988mental,Tabrez_2020}.

However, significant challenges arise with the implementation of ToM and mental model capabilities in handling the dynamic nature of human intentions in collaborative settings. Humans may change their approach mid-task or deviate from expected patterns for various reasons, requiring technical systems to distinguish between errors, intentional variations, and goal changes. This remains an active area of research.

Moreover, research has shown that humans are more likely to assign tasks to machines that they perceive as having mental capabilities, further emphasizing the importance of ToM in fostering effective collaboration \cite{Wiese_2022}. The ability for a machine to interpret the behaviors of another agent can help in determining security risks or creating more involved players in complex games. Thus, modeling ToM in machines enhances task performance and contributes to building trust and acceptance in human-machine interaction \cite{Lee2004TrustInAutomation}, while a lack thereof can lead to a breakdown in interaction \cite{Parasuraman1997HumansAndAutomation}.

\subsection{Human Error Classification}
\label{sssec:ECF}
To ground our model of behavioral deviations, we draw on Reason's Generic Error-Modeling System (GEMS), a framework that categorizes human errors based on their underlying cognitive processes \cite{Reason_1990}. GEMS distinguishes between skill-based errors like \emph{slips} (incorrect execution) and \emph{lapses} (memory failures); rule-based or knowledge-based \emph{mistakes} (flawed planning or judgment); and intentional \emph{violations} (e.g., deliberate procedural shortcuts) \cite{Reason_1990}.

In the context of human-machine interaction, these different error types manifest as observable deviations from expected task performance. Recognizing whether a deviation stems from an unintentional slip versus a deliberate violation is crucial for an agent to respond appropriately and ensure task success and safety. Therefore, the GEMS framework directly informs the design requirements for our ontology, providing a theoretical basis for classifying different types of deviations and their corresponding interpretations.

\section{Ontology Design Methodology}
\label{sec:meth}
Building on the foundational concepts described in the previous section, we now detail the specific engineering methodology used to construct OntoPret. This section outlines our process, which is grounded in the Modular Ontology Modeling (MOMO) framework and guided by the use of established ontology design patterns (ODPs) and competency questions (CQs).

\subsection{Ontology Modeling Approach}
\label{sub:ontoDesign}
Ontology is a specification of a conceptualization \cite{gruber}, meaning that an ontology acts as a description of various concepts via relationships that constrain how those concepts are related for a domain. OntoPret, grounded in ontology, utilizes the design of the MOMo \cite{momo-swj}. 
MOMo allows the design of the ontology to be reusable and adaptable as applications to use case scenarios and stakeholder needs evolve.
By using MOMo, OntoPret has adapted those characteristics of MOMo, increasing the applicability of the framework. As a byproduct, the MOMo methodology also enables the framework to integrate with other ontologies due to the emphasis on modularity, again increasing its usability and flexibility to adapt to future advancements. Furthermore,  the flexibility that the framework is built on plays a critical role in how the framework remains human-centric in its overall design. 

We particularly emphasized incorporating aspects from cognitive science. The multidisciplinary integration into the framework ensures a truly human-centric approach and that our vocabulary is consistent and appropriately sourced from cognitive science literature. This is in particular exemplified in our choice of labels for the Interpretation of Behavior, having either a \textsf{Confirmation} or a \textsf{Contradiction}, given specific expectations from a role. We have attempted to be thoughtful in our selection throughout the entire ontology, and we have significantly incorporated domain expert knowledge.

From the MOMo methodology, an emphasis was also placed on the reuse of applicable ODPs. ODPs are tiny, self-contained ontologies that model domain-invariant modeling problems \cite{odp-book}, and can be sourced from libraries (e.g., MODL \cite{modl}). These are then instantiated\footnote{This process is called \emph{template-based} instantiation and uses a pattern as a structural guide \cite{template}.} into modules, which are top-level conceptual components.\footnote{More details can be found in \cite{momo-swj} and how membership to a module can be understood (and annotated) in \cite{opla,opla-cp}.}  OntoPret makes direct use of the \textsf{AgentRole} design pattern to separate an Actor from the Role it performs. While other patterns were not formally instantiated, their core principles directly informed our design. Specifically, the \textsf{Tree} ODP guided our modeling of Goal hierarchies via the recursive hasSubGoal property. Likewise, the \textsf{Sequence} ODP provided the model for our Task structure, which uses hasNextStep and hasPreviousStep properties to represent ordered workflows.
However, this added in too much complexity. Finally, we considered using the explicit typing metapattern for delineating the various types but opted for directly modeling into the subsumption hierarchy without nominals. The ontology is formalized using OWL 2 DL \cite{owl-tr}, ensuring decidable reasoning. 

The resultant modular nature also facilitates the future incorporation or otherwise deeper integration of other cognitive ontologies that express human behaviors (e.g., HBO or BCIO).

\subsubsection{Application Context}
\label{ssec:apps}
During the concept development stage of OntoPret, the research team identified an initial key notion for the ontology: \emph{Behavior}, which set and shaped the direction of the ontology's development. The identification of other key notions for OntoPret was gathered through stakeholder consultations and domain expert input. These interactions provide additional perspectives on how the framework could be utilized,  highlighting more areas of interest. The final major key notions identified were:  \emph{Behavior}, \emph{Task-Allocation}, \emph{Environment}. 
These key notions provided a developmental checklist, guiding the creation of the CQs and, finally, the full ontology.

CQs are described as a set of questions formatted in natural language that the ontology must be able to answer correctly \cite{noy1997state}. CQs allow the ontology engineers to actively assess the accuracy, efficiency, and completeness of the ontology. CQs ensure that all important information is in the ontology and that the ontology is able to retrieve that information when prompted. The CQs developed for OntoPret followed the methodology outlined by Keet \& Khan (2024), which categorizes CQs into 5 main types: Scoped, Validate, Foundational, Metaproperties, and Relationship \cite{keet2024discerning}.

This methodology creates questions that build upon each other, ensuring that the ontology is reliable and contains the correct information. The identification of key notions and the potential stakeholder perspectives, when paired with this methodology, helps the ontology focus on the important aspects of its development and create CQs that support those aspects. Once the CQs were created, they were then converted into natural language. The conversion provides less complex documentation and helps users better comprehend the capabilities of the ontology. This conversion will also help with the integration of the ontology in a query language.

\begin{table}[t!]
\centering
\small
\caption{Selected competency questions for the Deviation and Deception scenarios.}
\label{tab:competency_questions}
\begin{tabularx}{\textwidth}{>{\bfseries}l|X|X}
\textbf{Area} & \textbf{Deviation Scenario Question} & \textbf{Deception Scenario Question} \\[1ex]
\hline\hline 
Behavior & ``\textit{If an Actor that performsRole AssemblerRole exhibits a Behavior that is an instance of Deviation, what is the resulting Interpretation?}'' & ``\textit{What are the top three Behaviors commonly exhibited by a person with a PlayerRole who is deceiving?}'' \\[2ex] \hline
Task Allocation & ``\textit{For a given Machine Actor, what Tasks is it specified to perform in order to achieve the Goal of the kitting Scenario?}'' & ''\textit{For a given person with a playerRole, what Tasks is it specified to perform to achieve the Goal of the poker Scenario?}'' \\[2ex] \hline
Environment & ``\textit{Within the Context of a Scenario, what properties or relations exist that link a Machine Actor to a Human Actor to represent situational awareness?}'' & ``\textit{For a set of Behavior instances classified as Deception, what is the average value of their hasResponseTime data property?}''
\end{tabularx}
\end{table}
 

\section{An Ontology for Interpreting Behavior}
\label{sec:ontology}
\begin{figure}[t]
  \begin{center}
    \includegraphics[width=\textwidth]{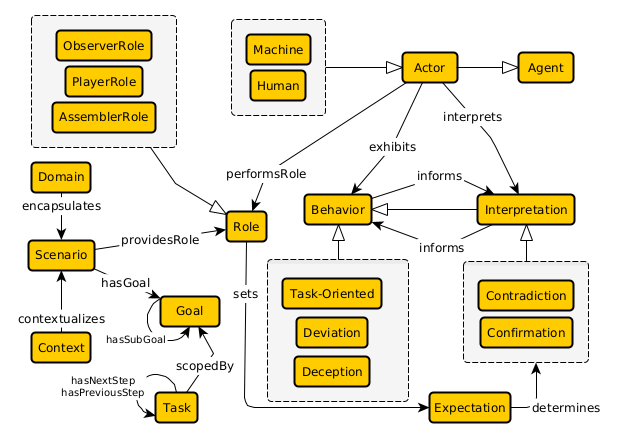}
  \end{center}
 \caption{Schema diagram for OntoPret. Gold boxes are classes, closed arrows indicate object properties, and open arrows indicate subsumption in the direction of the arrow. Grouped boxes are disjoint subclasses.}
 \label{fig: OntoPret}
\end{figure}
The Ontology for the Interpretation of Human Behavior (OntoPret) was developed to \emph{model} the various characteristics of how the behaviors of human agents in human-machine interaction environments are interpreted. Our goal is to provide an underlying ontological layer that enables safe, seamless, and effective interactions and collaborations in hybrid environments. A schema diagram for OntoPret is shown in Figure~\ref{fig: OntoPret}. The schema diagram may be interpreted as follows: Gold boxes are classes, closed arrows indicate object properties, open arrows indicate subsumption in the direction of the arrow, and grouped boxes are disjoint subclasses. \footnote{Note that edges in a schema diagram do not express cardinality and be minimally construed as representing the description logic axiom $\textsf{Head}\sqsubseteq\mathord{\geq}0\textsf{edge.Tail}$.} 

\begin{figure}[tppb]
    \centering
    \includegraphics[width=1\textwidth]{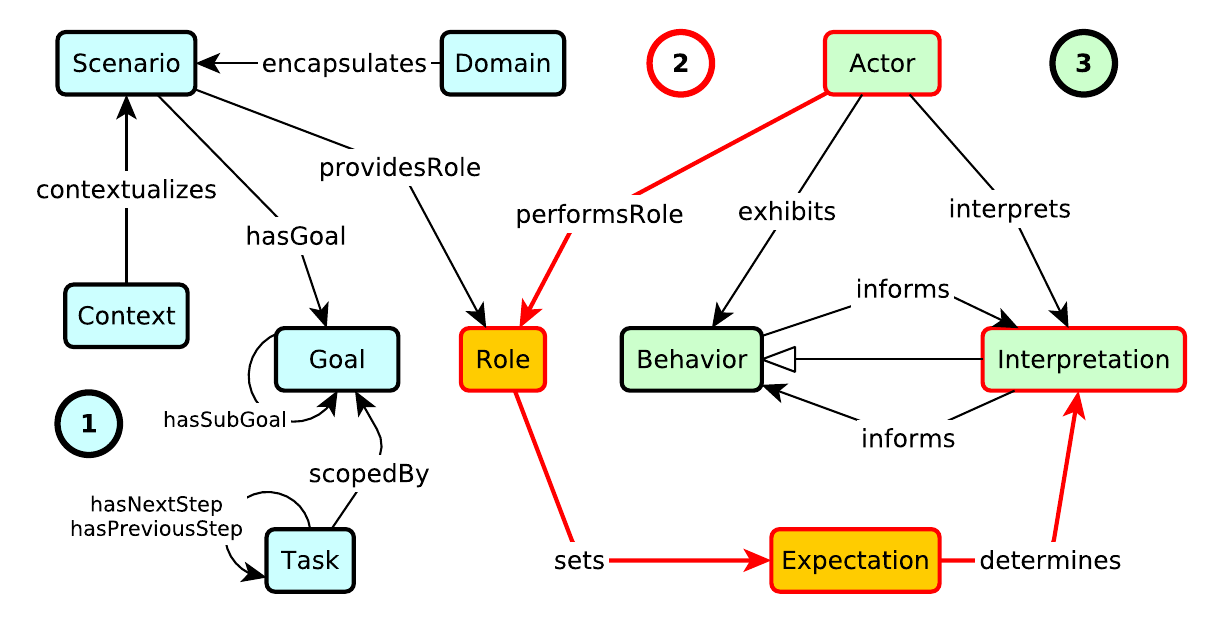}
     \caption{This figure shows OntoPret, but organized into three overarching conceptual components or modules: (1) the Scenario Module, (2) the Expectation Module, and (3) the Behavior Module.}
     \label{Figure:ontopret-modules}
\end{figure}

OntoPret can be broken down into three modules and is shown in Figure~\ref{Figure:ontopret-modules}: 
\begin{inparaenum}[(1)]
  \item the \emph{Scenario} Module, shown using the blue boxes;
  \item the \emph{Expectation} Module, shown using the red arrows and borders; and 
  \item the \emph{Behavior} Module, shown using the green boxes.
\end{inparaenum}  
Modules are the primary conceptual component in the MOMo methodology, and they are not necessarily disjoint groupings of concepts (and their relationships). As such, we see overlap particularly in the \textsf{Role} concept, and in \textsf{Actor} and \textsf{Interpretation}. These modules cooperate to inform and provide context to the interpretation of an agent's behavior. The scenario gives background to the role, the role sets the expectation, and the actor interprets and assesses the behavioral observations of the other agent(s) to their own interpretation determined by the expectation.
A primary design goal of OntoPret is to render complex human behavior in a formal, machine-processable manner. The framework achieves this by, first, classifying behaviors into discrete ontological classes (e.g., \textsf{Deviation}) and, second, defining logical axioms (in the form of axiom patterns \cite{owlaxax}) that govern how these classes are to be interpreted (e.g., that a \textsf{Deviation} must \textsf{informs} a \textsf{Contradiction}). The OWL files as well as the schemas can be found in the OntoPret Github Repository \cite{anonymous}. 

\paragraph{The Scenario Module}
OntoPret can be used in a variety of different Scenarios, exemplified by the use-cases in Section~\ref{sec:scenarios}. The \textsf{Scenario} concept is certainly complex, and it is best described in its components. We differentiate aspects of a scenario into its \textsf{Context} and its \textsf{Domain}. Essentially, the \textsf{Domain} represents an abstraction of the scenario. An example \textsf{Domain} might be \emph{manufacturing}. On the other hand, the \textsf{Context} acts as an ontological placeholder representing, for example, the environment (or spatiotemporal extent) in which the \textsf{Scenario} is occurring. As such, we note that the \textsf{encapsulates} and \textsf{conceptualizes} are both axiomatized as \emph{inverse existential}.\footnote{This axiom is scoped and qualified, and takes the form: $\textsf{Head}\sqsubseteq\exists\textsf{encapsulates}^-\textsf{.Tail}$.} A \textsf{Scenario} also has at least one \textsf{Goal}, which we consider to be an objective.\footnote{This is modeled as an existential axiom taking the form $\textsf{Head}\sqsubseteq\exists\textsf{edge.Tail}$.} \textsf{Goal}s furthermore, may have hierarchies of subgoals, and a \textsf{Goal} is achieved by completing \textsf{Task}s. In this, we differentiate between the objective (\textsf{Goal}) and the action taken to obtain that objective (\textsf{Task}).

Perhaps most importantly, the \textsf{Scenario} provides roles, and by this we mean in the sense of the \textsf{AgentRole} design pattern. In brief, this presents a separation between what an agent (or \textsf{Actor} in our case) \emph{does} and what it \emph{is}. The motivating purpose for an actor performing a role is dictated by the existence—and nature—of the scenario. For example, a poker game scenario might provide a role of a dealer, which defines expected behavioral patterns.

All together, we may take the \textsf{Scenario} to be a contextualized goal in some domain, where some actors must complete tasks to culminate the scenario. \textsf{Scenario}, and its surrounding topics stem from the work presented in \cite{tiddi2023knowledge}. Indeed, this would be a natural connection point.
From a reasoning perspective, this module enables a system to query the foundational context of an interaction, answering 'who, what, and why' questions (e.g., 'What is the \textsf{Goal} of the \textsf{Actor} performing this \textsf{Task}?')

\paragraph{The Expectation Module}
The \textsf{Expectation} module is essentially a chain of concepts that connects the Behavior module and the Scenario module. Specifically, we have an \textsf{Actor} that performs a \textsf{Role} (as provided by a \textsf{Scenario}) that sets \textsf{Expectation}s on how certain \textsf{Behaviors} can -- or should -- be interpreted (\textsf{Interpretation}). This enables context-sensitive behavioral assessment. Colloquially, we are essentially stating that ``context matters'' for the interpretation of an \textsf{Actor}'s behavior. Formally, we express these through existential axioms, meaning that all \textsf{Actor}s are performing a \textsf{Role}, whether or not it is known, and all \textsf{Role}s set \textsf{Expectation}s, which in turn always color or otherwise \textsf{determine} an \textsf{Interpretation}.

The \textsf{Expectation} is shaped through the role and, indirectly, through the scenario. As such, expectations (e.g., of behavior) may change between roles in the same scenario or even between identical roles in different scenarios. For example, we might consider the policing of different security measures in various transportation modalities. Speed is tightly regulated in an airspace, and thus expectations on violating these regulations differ significantly from speeding in a car on a highway or jaywalking through an empty intersection. We see that there are different understandings of \emph{baseline} or normality, which then go on to color the \textsf{Expectation} of an \textsf{Actor}. 


\paragraph{The Behavior Module}
The Actor-Behavior-\-Interpretation triad, is the human-centric component of OntoPret. 
\textsf{Behavior} is \textsf{exhibited} by an agent, which is determined by the role that they perform. For \textsf{Behavior}, we initially define three important subclasses. Our current understanding holds them to be mutually disjoint.
\begin{itemize}
    \item[] \textbf{Task-Oriented} -- This subclass is straightforward and self-explanatory. In this case, we would consider this to be the \emph{default} behavior of an \textsf{Actor} exhibited while completing a \textsf{Task}.
    \item[] \textbf{Deviation} -- This subclass identifies a \textsf{Contradiction} or a mismatch between an Actor's expected default behavior and their actual exhibited behavior. An example of deviant behavior is shown in Sec.~\ref{ssec:deviation}.
    \item[] \textbf{Deception} -- This subclass identifies behavior of a deceiving \textsf{Actor} that may cause the target \textsf{Actor} to believe something that they would not otherwise believe.
    An example of deceptive behavior is shown in Sec.~\ref{ssec:deception}.
\end{itemize}

For \textsf{Interpretation}, we initially define two important subclasses. Our current understanding also holds these to be mutually disjoint. (1) Contradiction: The expectations of an observer (or other Actor with a Role) have been somehow violated or mismatched against the observation. Specifically, this \emph{contradicts} their understanding and results in an unexpected behavior, which results in a label of deviant or deceptive behavior; and (2) Confirmation: An \textsf{Actor} will somehow interpret an observed behavior to conform to or otherwise match their expectations for a given Role and Scenario. This \emph{confirmation} of their expectation could possibly result in labeling the behavior as \textsf{Task-Oriented}.\footnote{This could also result in a \textsf{Deviant} behavior, as defined relative to achieving a task, but still may be expected (or predicted) by a sufficiently empowered system.}.

These conceptual classes are formally linked to create the core inferential capability of OntoPret. By leveraging the axioms, a reasoning engine can deduce the implications of an observed \textsf{Behavior}, such as concluding that a \textsf{Deviation} necessarily results in an \textsf{Interpretation} of \textsf{Contradiction}.The most critical of these are presented in Table \ref{tab:axioms}. 

\begin{table}[t!]
\centering
\renewcommand{\arraystretch}{1.2}
\caption{A selection of key axioms in OntoPret using Description Logic notation.}
\label{tab:axioms}
\begin{tabularx}{\textwidth}{l|X}
\textbf{Axiom} & \textbf{Description} \\[1ex]
\hline\hline

\texttt{Actor} $\sqsubseteq$ $\exists$\texttt{performsRole.Role} & Every Actor performs at least one Role. \\[1.5ex]
\texttt{Role} $\sqsubseteq$ $\exists$\texttt{sets.Expectation} & A Role always sets an Expectation. \\[1ex]

\hline
\texttt{Task-Oriented} $\sqsubseteq$ $\neg$\texttt{Deviation} & A Behavior cannot be both Task-Oriented and a Deviation. \\[1.5ex]
\texttt{Contradiction} $\sqsubseteq$ $\neg$\texttt{Confirmation} & An Interpretation cannot be both a Contradiction and a Confirmation. \\[1ex]

\hline
\texttt{Deviation} $\sqsubseteq$ $\forall$\texttt{informs.Contradiction} & A Deviation behavior only informs Interpretations that are Contradictions. \\[1.5ex]
\texttt{Task-Oriented} $\sqsubseteq$ $\forall$\texttt{informs.Confirmation} & A Task-Oriented behavior only informs Interpretations that are Confirmations.\\[1ex]
\end{tabularx}
\end{table}
\section{Use-Case Scenarios}
\label{sec:scenarios}
To demonstrate OntoPret's application, we present two distinct use-case scenarios: one involving \emph{deviation} in a collaborative manufacturing task, and a second involving \emph{deception} in a poker game. Both scenarios highlight the framework's flexibility, showing how its modular structure can be instantiated in different domains. Table~\ref{tab:scenarios_mapping} summarizes how the key concepts from each narrative map directly to the classes in OntoPret.

\begin{table}[t!]
\centering
\small 
\renewcommand{\arraystretch}{1.2}
\caption{OntoPret mappings to the Deviation and Deception scenarios.}
\label{tab:scenarios_mapping}
\begin{tabularx}{\textwidth}{l|X|X}

\textbf{Concept} & \textbf{Deviation (Manufacturing)} & \textbf{Deception (Poker)} \\[1ex]
\hline\hline
\textit{Scenario}   & Working in collaboration to correctly complete an assignment. & Playing a card-based game. \\ \hline
\textit{Context}    & Kitting task in an assembly line. & A poker game. \\ \hline
\textit{Domain}     & Manufacturing. & A betting game. \\ \hline
\textit{Goal}       & Correctly complete the kitting task. & Win the game and the pot of money. \\ \hline
\textit{Task}       & Retrieve specifically assigned items and place them on the table. & Make a choice to fold, call, or raise an opponent. \\ \hline
\textit{Role}       & Assembler (human), Observer (robot monitoring for deviations). & Player. \\ \hline
\textit{Agent/Actor} & Machine. & Machine. \\ \hline
\textit{Behavior}   & Human deviates from task by skipping to pick an assigned item. & Human displays deceptive/bluffing behavior by avoiding eye contact. \\ \hline
\textit{Interpretation} & A contradiction/confirmation of what the agent understands to be true to complete the kitting task. & A contradiction/confirmation that the player is holding a weak hand. \\ \hline
\textit{Expectation}  & Agents must correctly follow the specific instructions of the kitting task. & Players are trying to win the poker game; the player may be bluffing to cover a weak hand.
\end{tabularx}
\end{table}

\subsection{Scenario: Deviation}
\label{ssec:deviation}
To demonstrate a deviation scenario, we consider a collaborative kitting task on a manufacturing assembly line where a human worker and a robotic agent are jointly responsible for collecting components for production kits. The robot is equipped with a task model outlining the expected sequence of actions. Its role is to monitor the human's actions to ensure the kit is assembled correctly, inferring intent based on observed behavior.

During the kitting \textsf{Scenario}, the robot \textsf{Actor}, performing its \textsf{ObserverRole}, interprets the human's \textsf{Behavior}. When the human skips a bin, the robot classifies this \textsf{Behavior} as a \textsf{Deviation}. This is because the action \textsf{informs} an \textsf{Interpretation} of \textsf{Contradiction} against the \textsf{Expectation} set by the \textsf{AssemblerRole} and its associated \textsf{Task} sequence.

This deviation could stem from a number of causes. It might be a lapse - a momentary memory failure or distraction that led the human to forget the item. Alternatively, it could be a rule-based mistake, where the human mistakenly believed the item was not needed for this particular kit. In some cases, such deviations might even be deliberate violations, such as when a worker skips a step to save time under pressure.

Regardless of the cause, the robot must respond. It first verifies the deviation by cross-referencing the current state of the kit with the task model. Then, it reasons about the nature of the deviation: was it accidental or intentional? Depending on this assessment, the robot may choose to retrieve the missing item, prompt the human with a reminder, or escalate the issue if it poses a safety risk.

This scenario highlights the importance of deviation detection in collaborative robotics. By recognizing and responding to deviations - arising from either human error or intentional action - the robot helps ensure task success and also contributes to a safer, more resilient human-robot team.

\subsection{Scenario: Deception}
\label{ssec:deception}
To demonstrate the interpretation of deceptive behavior, we apply the framework to a poker game. In this scenario, communication is expected to involve bluffing, and a player's behavior (e.g., a nervous ``tell'') can be indicative of their hidden hand strength. The goal for a machine player is to interpret these potentially misleading behaviors to improve its assessment of opponents' hands and inform its decisions to fold, call, or raise.

The machine \textsf{Actor}, in its \textsf{PlayerRole}, continuously interprets the opponent's exhibited \textsf{Behavior}. A specific action, like avoiding eye contact, might be classified as an instance of \textsf{Deception}. This observed \textsf{Behavior} then \textsf{informs} the machine's \textsf{Interpretation}. If this aligns with the machine's assessment of the game state, it serves as a \textsf{Confirmation}, guiding the \textsf{Actor} to perform its next \textsf{Task} (e.g., to raise). If the \textsf{Behavior} is unexpected, it results in a \textsf{Contradiction}, prompting a different \textsf{Task} (e.g., to fold), all in service of the ultimate \textsf{Goal} of winning the game. This interpretation of the behavior sequence will continue until the round comes to an end and the winner is determined.

\section{Conclusions}
\label{sec:conc}
In this paper, we presented OntoPret, an ontology for the interpretation of human behavior. Our framework provides a human-centric approach to human-machine interaction by rooting its design in cognitive science and a modular engineering methodology. The resulting ontology is generalizable, flexible, and extendable, designed to provide a foundational tool that facilitates safer and more effective collaboration between humans and machines, as demonstrated in our use cases.

The primary \textbf{limitations} involve the need for more extensive validation and application. While OntoPret provides the formal structure, it requires domain-specific instantiation with a significant volume of data to be fully operational. Furthermore, a dedicated reasoning system must be integrated to fully leverage the ontological inferences in real-time. Finally, we have not yet performed a comprehensive structural evaluation against quantitative ontology metrics.

Building on this foundation, \textbf{future work} will proceed along several key paths. First, we will operationalize the framework by integrating OntoPret with intention recognition systems and formulating SPARQL queries to enable real-time behavioral reasoning and then validate these capabilities in simulated scenarios. We also plan to extend the ontology to support more nuanced, hierarchical modeling of intentions, allowing for reasoning over complex action sequences from micro-actions to strategic goals. Finally, we will investigate how OntoPret can be used for adaptive learning, using unexpected deviations and deceptive behaviors as signals to refine an agent's knowledge structures.

\bibliographystyle{splncs04}
\bibliography{refs, Alexis, Stacie}

\begin{thebibliography}{10}
\providecommand{\url}[1]{\texttt{#1}}
\providecommand{\urlprefix}{URL }
\providecommand{\doi}[1]{https://doi.org/#1}

\bibitem{anonymous}
Anonymous: Ontopret: Ontology for the interpretation of human behavior. \url{https://anonymous.4open.science/r/OntoPret-Ontology-for-the-Interpretation-of-Human-Behavior-4BE8/README.md} (2023), accessed: 2025-07-29

\bibitem{hadi}
Banaee, H., Kl{\"u}gl, F., Novakazi, F., Lowry, S.: Intention recognition and communication for human-robot collaboration. In: 3rd International Conference on Hybrid Human-Artificial Intelligence, HHAI-WS 2024, Malmo 10-11 June 2024. vol.~3825, pp. 101--108. CEUR-WS (2024)

\bibitem{beetz2018know}
Beetz, M., Be{\ss}ler, D., Haidu, A., Pomarlan, M., Bozcuo{\u{g}}lu, A.K., Bartels, G.: Know rob 2.0—a 2nd generation knowledge processing framework for cognition-enabled robotic agents. In: 2018 IEEE international conference on robotics and automation (ICRA). pp. 512--519. IEEE (2018)

\bibitem{carroll1988mental}
Carroll, J.M., Olson, J.R.: Mental models in human-computer interaction. Handbook of human-computer interaction pp. 45--65 (1988)

\bibitem{chisholm_intent_1977}
Chisholm, R.M., Feehan, T.D.: The {Intent} to {Deceive}. The Journal of Philosophy  \textbf{74}(3), ~143 (Mar 1977). \doi{10.2307/2025605}

\bibitem{cowork}
Demir, K.A., D{\"o}ven, G., Sezen, B.: Industry 5.0 and human-robot co-working. Procedia computer science  \textbf{158},  688--695 (2019)

\bibitem{valcri}
Dragisic, Z., Lambrix, P., Blomqvist, E.: Integrating ontology debugging and matching into the extreme design methodology. In: Blomqvist, E., Hitzler, P., Krisnadhi, A., Narock, T., Solanki, M. (eds.) Proceedings of the 6th Workshop on Ontology and Semantic Web Patterns {(WOP} 2015) co-located with the 14th International Semantic Web Conference {(ISWC} 2015), Bethlehem, Pensylvania, USA, October 11, 2015. {CEUR} Workshop Proceedings, vol.~1461. CEUR-WS.org (2015), \url{http://ceur-ws.org/Vol-1461/WOP2015\_paper\_1.pdf}

\bibitem{owlaxax}
Eberhart, A., Shimizu, C., Chowdhury, S., Sarker, M.K., Hitzler, P.: Most of {OWL} is rarely needed. In: 18th ESWC (2020), under review.

\bibitem{elwood_reading_2012}
Elwood, Z.: Reading {Poker} {Tells}. Via Regia (2012), \url{https://books.google.com/books?id=hLTnngEACAAJ}

\bibitem{dev3}
Endsley, M.R.: Toward a theory of situation awareness in dynamic systems. Human Factors  \textbf{37}(1),  32--64 (1995). \doi{10.1518/001872095779049543}

\bibitem{gentner2014mental}
Gentner, D., Stevens, A.L.: Mental models. Psychology Press (2014)

\bibitem{gruber}
Gruber, T.R.: A translation approach to portable ontology specifications. Knowledge acquisition  \textbf{5}(2),  199--220 (1993)

\bibitem{guazzini_emergence_2009}
Guazzini, A., Vilone, D.: The emergence of bluff in poker-like games (2009), \url{https://arxiv.org/abs/0901.3365}

\bibitem{template}
Hammar, K., Presutti, V.: Template-based content {ODP} instantiation. In: et~al., K.H. (ed.) Advances in Ontology Design and Patterns [revised and extended versions of the papers presented at the 7th edition of the Workshop on Ontology and Semantic Web Patterns, WOP@ISWC 2016, Kobe, Japan, 18th October 2016]. Studies on the Semantic Web, vol.~32, pp. 1--13. {IOS} Press (2016). \doi{10.3233/978-1-61499-826-6-1}

\bibitem{opla-cp}
Hirt, Q., Shimizu, C., Hitzler, P.: Extensions to the ontology design pattern representation language. In: Janowicz, K., Krisnadhi, A.A., Poveda{-}Villal{\'{o}}n, M., Hammar, K., Shimizu, C. (eds.) Proceedings of the 10th Workshop on Ontology Design and Patterns {(WOP} 2019) co-located with 18th International Semantic Web Conference {(ISWC} 2019), Auckland, New Zealand, October 27, 2019. {CEUR} Workshop Proceedings, vol.~2459, pp. 76--75. CEUR-WS.org (2019), \url{http://ceur-ws.org/Vol-2459/short2.pdf}

\bibitem{odp-book}
Hitzler, P., Gangemi, A., Janowicz, K., Krisnadhi, A., Presutti, V. (eds.): Ontology Engineering with Ontology Design Patterns: Foundations and Applications, Studies on the Semantic Web, vol.~25. IOS Press, Amsterdam (2016)

\bibitem{opla}
Hitzler, P., Gangemi, A., Janowicz, K., Krisnadhi, A.A., Presutti, V.: Towards a simple but useful ontology design pattern representation language. In: Blomqvist, E., Corcho, {\'{O}}., Horridge, M., Carral, D., Hoekstra, R. (eds.) Proceedings of the 8th Workshop on Ontology Design and Patterns {(WOP} 2017) co-located with the 16th International Semantic Web Conference {(ISWC} 2017), Vienna, Austria, October 21, 2017. {CEUR} Workshop Proceedings, vol.~2043. CEUR-WS.org (2017), \url{http://ceur-ws.org/Vol-2043/paper-09.pdf}

\bibitem{owl-tr}
Hitzler, P., Parsia, B., Rudolph, S., Patel-Schneider, P., Kr{\"{o}}tzsch, M.: {OWL} 2 web ontology language primer (second edition). {W3C} recommendation, W3C (Dec 2012), https://www.w3.org/TR/2012/REC-owl2-primer-20121211/

\bibitem{dev1}
Hoffman, G., Breazeal, C.: Effects of anticipatory action on human-robot teamwork. In: Proceedings of the ACM/IEEE International Conference on Human-Robot Interaction (HRI). pp.~1--8. ACM (2007). \doi{10.1145/1228716.1228717}

\bibitem{dev2}
{International Organization for Standardization}: {ISO/TS 15066:2016 Robots and robotic devices — Collaborative robots}. \textit{International Organization for Standardization} (2016), \url{https://www.iso.org/standard/62996.html}

\bibitem{keet2024discerning}
Keet, C.M., Khan, Z.C.: Discerning and characterising types of competency questions for ontologies. arXiv preprint arXiv:2412.13688  (2024)

\bibitem{king_applications_2024}
King, S., Neal, T.: Applications of {AI}-{Enabled} {Deception} {Detection} {Using} {Video}, {Audio}, and {Physiological} {Data}: {A} {Systematic} {Review}. IEEE Access  \textbf{PP}, ~1--1 (Jan 2024). \doi{10.1109/ACCESS.2024.3462825}

\bibitem{aiindustrial}
Lee, J., Davari, H., Singh, J., Pandhare, V.: Industrial artificial intelligence for industry 4.0-based manufacturing systems. Manufacturing letters  \textbf{18},  20--23 (2018)

\bibitem{Lee2004TrustInAutomation}
Lee, J.D., See, K.A.: Trust in automation: Designing for appropriate reliance. Human Factors  \textbf{46}(1),  50--80 (2004). \doi{10.1177/001872080404600104}

\bibitem{lemaignan2010oro}
Lemaignan, S., Ros, R., M{\"o}senlechner, L., Alami, R., Beetz, M.: Oro, a knowledge management platform for cognitive architectures in robotics. In: 2010 IEEE/RSJ International conference on intelligent robots and systems. pp. 3548--3553. IEEE (2010)

\bibitem{Li_2024}
Li, W., Hu, Y., Zhou, Y., Pham, D.T.: Safe human–robot collaboration for industrial settings: a survey. Journal of Intelligent Manufacturing  (2024)

\bibitem{mahon_definition_2015}
Mahon, J.: The {Definition} of {Lying} and {Deception}. Stanford Encyclopedia of Philosophy  (Dec 2015)

\bibitem{marques2024behaviour}
Marques, M.M., Wright, A.J., Corker, E., Johnston, M., West, R., Hastings, J., Zhang, L., Michie, S.: The behaviour change technique ontology: transforming the behaviour change technique taxonomy v1. Wellcome open research  \textbf{8}, ~308 (2024)

\bibitem{michie2021representation}
Michie, S., West, R., Finnerty, A.N., Norris, E., Wright, A.J., Marques, M.M., Johnston, M., Kelly, M.P., Thomas, J., Hastings, J.: Representation of behaviour change interventions and their evaluation: Development of the upper level of the behaviour change intervention ontology. Wellcome open research  \textbf{5}, ~123 (2021)

\bibitem{Nahavandi2019Industry50}
Nahavandi, S.: Industry 5.0—a human-centric solution. Sustainability  \textbf{11}(16), ~4371 (2019). \doi{10.3390/su11164371}

\bibitem{noy1997state}
Noy, N.F., Hafner, C.D.: The state of the art in ontology design: A survey and comparative review. AI magazine  \textbf{18}(3),  53--53 (1997)

\bibitem{Ottoni_2024}
Ottoni, L.T.C., Cerqueira, J.J.F.: A systematic review of human–robot interaction: The use of emotions and the evaluation of their performance. International Journal of Social Robotics  (2024)

\bibitem{Parasuraman1997HumansAndAutomation}
Parasuraman, R., Riley, V.: Humans and automation: Use, misuse, disuse, abuse. Human Factors  \textbf{39}(2),  230--253 (1997). \doi{10.1177/001872089703900204}

\bibitem{Rahman2024Cobotics}
Rahman, M.M., Khatun, F., Jahan, I., Devnath, R., Al-Amin, M.: Cobotics: The evolving roles and prospects of next-generation collaborative robots in industry 5.0. Journal of Robotics  (2024)

\bibitem{Reason_1990}
Reason, J.: Human Error. Cambridge University Press (1990), this is the foundational source for GEMS and the SRK model.

\bibitem{sarkadi_deception_2024}
Sarkadi, S.: Deception {Analysis} with {Artificial} {Intelligence}: {An} {Interdisciplinary} {Perspective} (Jun 2024). \doi{10.48550/arXiv.2406.05724}, \url{http://arxiv.org/abs/2406.05724}, arXiv:2406.05724 [cs]

\bibitem{human}
Schenk, P.M., West, R., Castro, O., et~al.: An ontological framework for organising and describing behaviours: The human behaviour ontology. Wellcome Open Research  \textbf{9}, ~237 (2024). \doi{10.12688/wellcomeopenres.21252.1}, \url{https://doi.org/10.12688/wellcomeopenres.21252.1}

\bibitem{industry4.0}
Schwab, K.: The fourth industrial revolution. https://www.weforum.org/about/the-fourth-industrial-revolution-by-klaus-schwab/ (2016)

\bibitem{momo-swj}
Shimizu, C., Hammar, K., Hitzler, P.: Modular ontology modeling. Semantic  (2021), in Press

\bibitem{modl}
Shimizu, C., Hirt, Q., Hitzler, P.: {MODL:} {A} modular ontology design library. In: Janowicz, K., Krisnadhi, A.A., Poveda{-}Villal{\'{o}}n, M., Hammar, K., Shimizu, C. (eds.) Proceedings of the 10th Workshop on Ontology Design and Patterns {(WOP} 2019) co-located with 18th International Semantic Web Conference {(ISWC} 2019), Auckland, New Zealand, October 27, 2019. {CEUR} Workshop Proceedings, vol.~2459, pp. 47--58. CEUR-WS.org (2019), \url{http://ceur-ws.org/Vol-2459/paper4.pdf}

\bibitem{smartrobotic}
Soori, M., Dastres, R., Arezoo, B., Jough, F.K.G.: Intelligent robotic systems in industry 4.0: A review. Journal of Advanced Manufacturing Science and Technology pp. 2024007--0 (2024)

\bibitem{Tabrez_2020}
Tabrez, A., Luebbers, M.B., Hayes, B.: A survey of mental modeling techniques in human–robot teaming. Current Robotics Reports  (2020), theory of mind in HRC

\bibitem{tiddi2023knowledge}
Tiddi, I., De~Boer, V., Schlobach, S., Meyer-Vitali, A.: Knowledge engineering for hybrid intelligence. In: Proceedings of the 12th Knowledge Capture Conference 2023. pp. 75--82 (2023)

\bibitem{umbrico2020ontology}
Umbrico, A., Orlandini, A., Cesta, A.: An ontology for human-robot collaboration. Procedia CIRP  \textbf{93},  1097--1102 (2020)

\bibitem{westbrook2006mental}
Westbrook, L.: Mental models: a theoretical overview and preliminary study. Journal of Information Science  \textbf{32}(6),  563--579 (2006)

\bibitem{Wiese_2022}
Wiese, E., Weis, P.P., Bigman, Y., Kapsaskis, K., Gray, K.: It’s a match: Task assignment in human–robot collaboration depends on mind perception. International Journal of Social Robotics  (2022). \doi{10.1007/s12369-021-00789-0}

\bibitem{5.0industry1}
Xu, X., Lu, Y., Vogel-Heuser, B., Wang, L.: Industry 4.0 and industry 5.0—inception, conception and perception. Journal of manufacturing systems  \textbf{61},  530--535 (2021)

\end{thebibliography}
\end{document}